\pdfoutput=1
\documentclass[11pt]{article}

\usepackage{amsmath,amsfonts,bm}

\def\eqref#1{equation~\ref{#1}}
\def\1{\bm{1}}

\DeclareMathAlphabet{\mathsfit}{\encodingdefault}{\sfdefault}{m}{sl}
\SetMathAlphabet{\mathsfit}{bold}{\encodingdefault}{\sfdefault}{bx}{n}

\usepackage[]{acl}

\usepackage{times}  % DO NOT CHANGE THIS
\usepackage{helvet}  % DO NOT CHANGE THIS
\usepackage{courier}  % DO NOT CHANGE THIS
\usepackage[utf8]{inputenc} % allow utf-8 input
\usepackage{microtype}
\usepackage{subfigure}
\usepackage{comment}
\usepackage{mathtools}
\usepackage{enumitem}
\usepackage{multicol,multirow}
\usepackage{arydshln}
\usepackage{makecell}
\usepackage{array}
\usepackage{bm}
\usepackage{cancel}
\usepackage{parskip}
\usepackage[T1]{fontenc}    % use 8-bit T1 fonts
\usepackage{booktabs}       % professional-quality tables
\usepackage{amsmath, amsthm}
\usepackage{amssymb}
\usepackage{amsfonts}
\usepackage{pifont}
\usepackage{hyperref}

\usepackage{algorithm}
\usepackage{algorithmic}

\newenvironment{tightitemize}%
  {\begin{itemize}[topsep=0pt, partopsep=0pt] %
    \setlength{\itemsep}{0pt}%
    \setlength{\parskip}{0pt}%
    }%
  {\end{itemize}}

\title{Summarize, Outline, and Elaborate: Long-Text Generation via Hierarchical Supervision from Extractive Summaries}

\author{
Xiaofei Sun$^{1,2}$, Zijun Sun$^2$, Yuxian Meng$^2$,
 Jiwei Li$^{1,2}$ and Chun Fan$^{3}$\\
  $^1$Zhejiang University, $^2$Shannon.AI, $^3$Computer Center, Peking University \\
  $^3$National Biomedical Imaging Center, Peking University, 
  $^3$Peng Cheng Laboratory \\
   xiaofei\_sun@zju.edu.cn, jiwei\_li@shannonai.com
}

\begin{document}
\maketitle
\begin{abstract}
The difficulty of generating coherent long texts lies in the fact that existing  models overwhelmingly focus on predicting local words,  and cannot make high level plans on what to generate or capture the high-level discourse dependencies between chunks of texts. 
Inspired by human writing processes, where a list of bullet points or a catalog is first outlined, and then each bullet point is  expanded to form the whole article, we propose {\it SOE}, a pipelined system that involves of summarizing, outlining and elaborating for long text generation: the model first outlines the summaries for different segments of long texts, and then elaborates on each bullet point to generate the corresponding segment. To avoid the  labor-intensive process of summary soliciting, we propose the {\it reconstruction} strategy, which extracts segment summaries in an unsupervised manner by selecting its most informative part to reconstruct the segment. 

The proposed generation system  comes with the following merits: (1) the summary provides  high-level guidance for text generation and avoids  the local minimum of individual word predictions; (2) the high-level discourse dependencies are captured in the conditional dependencies between summaries and are preserved during the summary expansion process and (3) additionally, we are able to consider significantly more contexts by representing contexts as concise summaries. Extensive experiments demonstrate that SOE produces long texts with significantly better quality, along with faster convergence speed. 
\footnote{Accepted by COLING 2022.}
\end{abstract}

\section{Introduction}
%Over recent years, large-scale pretrained language models (PLMs) have made enormous achievements across a wide range of natural language processing tasks such as natural language understanding \citep{elmo,devlin2018bert,yinhan2019roberta,yang2019xlnet,clark2020electra}, machine translation \citep{lample2019cross,liu2020multilingual}, information extraction \citep{danqi2020spanbert}, question answering \citep{karpukhin2020dense} and natural language generation \citep{dai-etal-2019-transformer,song2019mass,dong2019unified,zhang2019pegasus,lewis2019bart}. These PLMs have particularly benefited text generation, leading to more fluent, coherent and informative text \citep{radford2019gpt2,li2020optimus,brown2020language}.

Despite that recent large-scale pretrained language models (PLMs) \citep{devlin2018bert,yinhan2019roberta,yang2019xlnet} are able to produce high-quality passages that can be hardly recognized by humans \citep{zellers2019defending}, 
most of the generated ``good'' texts are within very limited length, e.g. hundreds of tokens for most cases \citep{yan2020prophetnet}, thus generating coherent long texts remains a challenge \cite{radford2019gpt2,tan2020progressive}. 
The difficulty lies in the fact that existing models generate texts in a  word-by-word manner: predicting each  subsequent token given its preceding contexts.
%using the softmax objective. 
This word-by-word strategy overwhelmingly focuses on predicting local words, and cannot  make high level plans on what to generate. The strategy results in the fact that long texts generated by current models are usually repetitive, generic and self-contradictory~\citep{shen2019towards}. 

To address these issues, the coarse-to-fine generation strategy  is proposed \cite{fan2018hierarchical,xu2018skeleton,yao2019planandwrite,mao2019improving}. 
In coarse-to-fine generation, a list of keywords or a short prompt is first generated, serving as a summary of the original text.
The prompt is then fed to a seq2seq model as an input to output the complete text. The coarse-to-fine generation strategy significantly improves generation over the word-by-word strategy, but still suffers from the following shortcomings: 
(a) {\it limited capacity of the prompt}: 
a single keyword list or prompt does not have enough capacity to summarize all the text of long passages, since long texts are usually consists of several parts, each of which focuses on a  specific aspect  or topic \citep{zhou-etal-2018-neural-storyline,guan2019story}.
The usage of the coarse-to-fine generation strategy is thus  limited to texts that can be summarized by a single prompt (e.g., short stories). 
This explains why text length generated by the progressive generation model is still limited, e.g., the introduced writing prompts dataset in \citet{fan2018hierarchical} has an average length of stories around 735, and the average length of prompts is 28;
 (b) {\it ignorance of high-level discourse dependency}:
 the coarse-to-fine generation strategy does not capture discourse-level dependencies \cite{li2016neural}, which handle the high-level information flow and interactions between  segments of texts. 
The ignorance of discourse-level dependencies results in texts lacking for coherence. 

Humans write in a hierarchical top-down manner: 
before writing a thousand-word-long essay,  
a human usually first prepares a list of bullet points or a catalogue, and then expands them to form the whole article. 
The sentence-level coherence between these bullet points  is preserved when the bullet points are expanded, 
providing guarantees that the full text is coherent. 

To mimic this top-down manner in human writing, 
we propose {\it SOE}, a pipelined system that involves of summarizing, outlining and expanding for long text generation: 
the model first outlines the summaries for different segments of long texts, which actually mimics the process of humans outlining bullet points;
next, the model elaborates on each bullet point to generate the corresponding segment. 
The proposed strategy  comes with the following merits:
(a) Since each segment is associated with its own summary rather than the entire text sharing a single prompt, the  capacity of summaries to reconstruct the full text can be guaranteed; 
(b) The conditional generation probability between 
 summaries  captures the high-level discourse dependencies, and these dependencies are preserved when they are expanded to segments. 
 This
 naturally resolves the incapability of modeling discourse-level dependencies in the coarse-to-fine generation approach. 
(c) This model is able to consider  significantly larger amount of contexts by representing chunks of contexts as concise summaries.

Empirically, we do not readily have summaries for segments in hand.  
The model thus needs to learn to summarize in an unsupervised manner. To this end, we propose the {\it reconstruction} strategy, which extracts segment summaries by selecting its most informative part to reconstruct the segment. 
 Extensive experiments demonstrate that SOE produces long texts with significantly better quality than existing baselines.

\section{Related Work}
\label{related}
\subsection{Generating Long Texts}
There are two trends for generating long text:
This first trend of work tackles the problem 
 from the model perspective. New  model structures \cite{kitaev2020reformer,child2019generating,dai-etal-2019-transformer,ye2019bp,guo-etal-2019-star,sukhbaatar-etal-2019-adaptive,correia-etal-2019-adaptively,beltagy2020longformer,zaheer2020big} are designed to give the model the ability to congest more contexts given limited memories or computing power. 
 For example, Transformer-XL \citep{dai-etal-2019-transformer},  a modifier to Transformers \citep{vaswani2017transformer}
  uses a segment-level recurrence mechanism 
 to enable learning long-term dependencies;
 \citet{child2019generating,correia-etal-2019-adaptively,kitaev2020reformer,beltagy2020longformer,zaheer2020big} proposed to sparsify transformers by focusing only on a fraction of attention connections;
 \citet{tay2020synthesizer} replaced the dot-product self-attention with learned synthetic attention weights.

The second trend of researches  focus on developing new generation strategies. 
 Efforts have been devoted to the idea of planning-then-generation
 or coarse-to-fine  generation
  \citep{sha2017order,gehrmann-etal-2018-end,wiseman2019learning,hua2019sentencelevel,shen2020blank,fu2020paraphrase}, which greatly inspires this work. 
  In coarse-to-fine generation, a list of keywords or a short sentence is first generated, providing guidance to generate the full text. 
%  Our model is greatly inspired by recent efforts in the progressive approach from \citet{fan2018hierarchical,liu2018generating}.
%Take story generation as an example,  decomposed the generation process by first generating the premise or prompt of the story and then generating the full story conditioning on the premise.
A recent work from \citet{tan2020progressive} takes a multi-step strategy, 
which progressively refines the generated incomplete text until reaching a specified stage.
Similar ideas have also been applied to text summarization, where \citet{gehrmann2018bottom} proposed a bottom-up method that first identifies phrases within a document that are likely included in its summary.
Our work is also inspired by the strategy of hierarchical  generation \cite{li2015hierarchical}, which consider text units with bigger granularity: 
 \citet{li2015hierarchical} proposed 
  hierarchical LSTMs that arrange tokens, sentences and paragraphs
in a hierarchical structure, with different levels of
LSTMs capturing compositionality, and 
\citet{shen2019towards} used 
 multi-level structures to learn
a VAE model for generating long coherent text.

%One recent work \citep{tan2020progressive} adopts a similar way to ours, . Although we both iteratively generate text, we argue that there are three key differences between ours and \citet{tan2020progressive}: 
%(1) their method is based on the full text while our proposed method only works on one segment at a time; 
%(2) the generated partial text by the method of \citet{tan2020progressive} is usually not continuous in the original text (see Figure 2 in \citet{tan2020progressive}), while our method follows a segment-by-segment autoregressive way; 
%and (3) their method needs to constrain the vocabulary at each stage, while ours simply uses the same full vocabulary for all segments. 

\subsection{Extractive Summarization}
Extractive summarization refers to the problem of selecting part of the input text as its summary. 
A fundamental problem in extractive summarization is to score constituent texts units (e.g., phrases, sentences or paragraphs) and select highly-ranked one(s) as the summary. 
\citet{haghighi2009exploring} used word frequencies in the input text  to assign scores to words, which are then in turn used to score sentences. 
Higher-ranked sentences are selected as the summary of the input text. 
\citet{liu2018generating} presented a two-stage extractive-abstractive framework, which first coarsely identifies salient information, followed by a generation model used to refine it.
Neural models have been widely used for scoring \citet{cao2015ranking,ren2017leveraging}.
\citet{liu2019text} finetuned BERT \citep{devlin2018bert} to score each sentence for extractive summarization; 
\citet{zhang2019bertscore} computed token similarity in each sentence using BERT contextual embeddings to serve as an automatic evaluation metric for text generation.

\section{Background}
\label{background}
\paragraph{Language Modeling}
refers to the process of calculating the probability $p(\mathbf{y})$ of a sequence
$\mathbf{y}=[y_1,\cdots,y_T]$, where each $y_i$ denotes a constituent token of $p(\mathbf{y})$. 
The probability can be computed by decomposing the joint distribution $p(\mathbf{y})$ into a product of conditional distributions over tokens:
\begin{equation}
  p(\mathbf{y})=\prod_{t=1}^Tp(y_t|\mathbf{y}_{<t})
\end{equation}
where $\mathbf{y}_{<i}=[y_1,\cdots,y_{i-1}]$ is the partial sequence of tokens generated previously. During training, the model is optimized to minimize the negative log-likelihood (NLL) $-\sum_{\mathbf{y}\in\mathcal{D}}\log p(\mathbf{y})$. During inference, the model decodes a token at each time step $t$ according to $p(y_t|\mathbf{y}_{<t})$ based on the softmax functions $y_t\propto \text{softmax}(\mathbf{W}^\text{out}\mathbf{h}_{t})$ where $\mathbf{W}^\text{out}\in\mathbb{R}^{d\times |V|}$ is the output word embedding matrix and $\mathbf{h}_{t}$ is the hidden state at time-step $t$.

\paragraph{Sequence-to-Sequence (Seq2Seq) Generation}
 models generate a target sequence $\mathbf{y}$ conditioning on a given source sequence $\mathbf{x}$, which differs from language models (LMs) in terms of whether or not conditioning on another input sequence. Similar to LMs, the probability of the target sequence can be typically factorized as:
\begin{equation}
  p({\mathbf{y}}|\mathbf{x})=\prod_{i=1}^Tp({y}_t|\mathbf{y}_{<t},\mathbf{x})
\end{equation}
Seq2seq models are also optimized to minimize the NLL $-\sum_{(\mathbf{x},{\mathbf{y}})\in\mathcal{D}}\log p({\mathbf{y}}|\mathbf{x})$. In the rest of this paper, we unify the notation of $p(\mathbf{y})$ and $p({\mathbf{y}}|\mathbf{x})$ by setting $\mathbf{x}=\varnothing$ for LMs.
Different architectures have been proposed to model $p({y}_t|{\mathbf{y}}_{<t},\mathbf{x})$, including transformers \citep{vaswani2017transformer}, 
LSTMs \citep{luong2015effective} and CNNs \citep{dauphin2017language}. 
At test time, sequences are usually generated 
using beam search, or its variants to promote diversity  \citep{vijayakumar2016diverse}.  

\section{Model Details for SOE}
\label{model}
\subsection{Notations}
A long sequence of tokens $\mathbf{Y} = \{\mathbf{y}^1, \mathbf{y}^2, \cdots, \mathbf{y}^K\}$ is first sliced into a series of snippets $\mathbf{y}^i$s, where $K$ denotes the number of constituent snippets.
Here we use the bold font $\mathbf{y}$ to denote snippets, and the normal font $y$ to denote tokens.
The number of tokens $N$ within each snippet is a hyper-parameter. 
 We also use
superscript $i$ to denote the index of a snippet, 
and subscript $l$ to denote the index of a token. 
 Each $\mathbf{y}^i$ consists a sequence of tokens $\mathbf{y}^i = \{y^i_1,\cdots,y^i_{n_i}\}$, where $n_i$ denotes the length of $\mathbf{y}^i$. 
Our goal is to generate 
a subset of $\mathbf{Y}$, denoted by $\mathbf{y}^{j\sim k} = \{\mathbf{y}^j,\mathbf{y}^{j+1},\cdots,\mathbf{y}^k\}$
given its preceding snippets, denoted by $p(\mathbf{y}^{j\sim k}  | \mathbf{y}^{<j})$.
Each snippet $\mathbf{y}^i$ is associated with a short summary $\mathbf{s}^i = \{s^i_1, s^i_2, \cdots, s^i_{m_i}\}$,
where $s^i_l$ denotes tokens and $m_i$ is the number of tokens in $\mathbf{s}^i$.

\begin{figure}[t]
  \centering
  \includegraphics[scale=0.4]{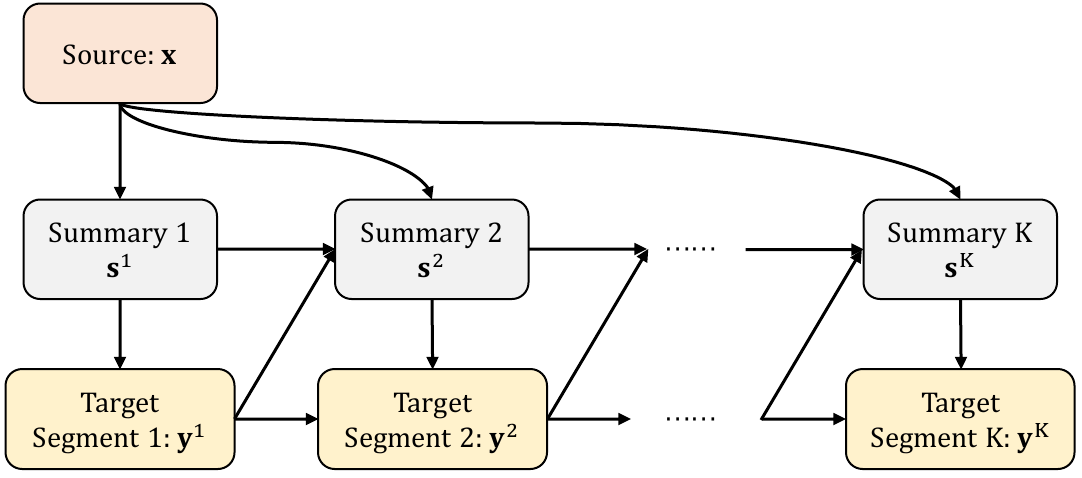}
  \caption{An overview of the proposed method. Given preceding tokens $\mathbf{y}^{<i}$, we first sequentially generate summaries   $\mathbf{s}^{j\sim k}$ for each snippet. Next we expand each summary $\mathbf{s}$ to form the full text  $\mathbf{y}^{j\sim k}$ .}
  \label{fig:illustration}
\end{figure}

\subsection{Pipeline Overview}
Instead of generating all constituent words in $\mathbf{Y}$ one by one, we adopt a hierarchical strategy. 
The  process of generating $\mathbf{y}^{j\sim k}$ is decoupled into the following two stages. 

{\bf Stage1: Outlining Segment Summaries }:
we sequentially generate the summary $\mathbf{s}^i$ for each snippet conditioning on the summaries for previous snippets. 
This mimics the process of catalog generation when humans write. 

{\bf Stage2: Expanding Summaries to Texts}: 
we expand each summary $ \mathbf{s}^i$  to the full segment by sequentially generating its constituent words.

\noindent An overview of the proposed method is shown in Figure \ref{fig:illustration}. 
\subsection{Extracting Golden Summaries}
\label{extract}
At the training time, we need to learn to generate summaries. 
But this is not straightforward because the golden summary 
 $\mathbf{s}^i$ for the snippet $\mathbf{y}^i$ is not readily at hand. 
Manually soliciting summaries like \citet{fan2018hierarchical} is both costly and slow. 
We thus propose to take the idea of unsupervised extractive summarization, and for each snippet $\mathbf{y}^i$, we extract its summary  $\mathbf{s}^i$ unsupervisedly,
and use the extracted $\mathbf{s}^i$ as the golden summary for learning.

We investigate Random, TF-IDF~\citep{Ramos2003UsingTT}, TextRank~\citep{mihalcea2004textrank} and Reconstruction methods to access the importance of selecting summary sentences, the first three of which are similar to \citet{liu2018generating}. More details of these three methods can be found at \ref{sec:appendix0}. 

\begin{comment}
 \paragraph{Random} For comparing purposes, we use a random sentence as the summary.
 \paragraph{TF-IDF} We take the sentence with the highest average TF-IDF score \citep{Ramos2003UsingTT} as the golden summary ${\mathbf{s}}^i$. A word is assigned a score by TF-IDF that scales proportionally to the number of times the word appears in the document and is offset by the number of documents in the corpus that contain the word, which can be expressed as $N_w\cdot\log(\frac{N_d}{N_{dw}})$, where $N_w$ is the word count, $N_d$ is the total number of documents and $N_{dw}$ is the total number of documents containing the word.
 \paragraph{TextRank} TextRank \citep{mihalcea2004textrank} is a weighted graph with tokens as nodes and the similarity between nodes as edges. We use BERT \citep{devlin2018bert} to compute the similarities between sentences and then rank them based on the TextRank algorithm. 
\end{comment}
 
 \paragraph{Reconstruction} A summary should be more informative than non-summary sentences, that is, a summary should have the most ability to reconstruct the full text. 
 To measure the degree of a sentence's {\it reconstruction} ability, we use a seq2seq model to predict 
 the original given text the summary sentence, the probability of which is regarded as the reconstruction score. Suppose there are $n$ sentences in $\mathbf{y}^i$ (e.g., $n=4$), and $\mathbf{y}^i=\{\mathbf{y}^{i,1}, \mathbf{y}^{i,2}, \mathbf{y}^{i,3}, \mathbf{y}^{i,4}\}$, 
 and $ \mathbf{y}^{i,j}$ denotes the $j$-{th} sentence in $\mathbf{y}^i$.
 The reconstruction score for $\mathbf{y}^{i,j}$, denoted by $\text{Score}(\mathbf{y}^{i,j})$ is given as follows:
 \begin{equation}
 \text{Score}(\mathbf{y}^{i,j}) = \frac{1}{|\mathbf{y}^i|} \log p(\mathbf{y}^i|\mathbf{y}^{i,j})
 \end{equation}
 To obtain $p(\mathbf{y}^i|\mathbf{y}^{i,j})$, we train another seq2seq model, where the input is $\mathbf{y}^{i,j}$ for each $j$, and the output is $\mathbf{y}^i$ by sequentially predicting tokens in $\mathbf{y}^i$. 
Given the trained model, we rank all sentences in  $\mathbf{y}^i$ 
and use the one with the highest score as the 
  golden summary ${\mathbf{s}}^i$.

\subsection{Outlining Segment Summaries}
In the summary generation stage, we cannot observe $\mathbf{y}^{j\sim k}$,
and our goal is to sequentially generate $\mathbf{s}^{j\sim k}$ given  $\mathbf{y}^{<j}$:
\begin{equation}
p( \mathbf{s}^{j\sim k}|\mathbf{y}^{<i}) = \prod_{i\in [j,k]} p(\mathbf{s}^i |\mathbf{y}^{<i}, \mathbf{s}^{<i}) \\
\label{generate-summary}
\end{equation}
The generation of summary $\mathbf{s}^i$ can be factorized into  sequentially generating the constituent word within it:
\begin{equation}
p(\mathbf{s}^i |\mathbf{y}^{<i}, \mathbf{s}^{<i}) = \prod_{l\in [1, m_i]} p(s^i_l | {s}^i_{<l}, \mathbf{y}^{<i}, \mathbf{s}^{<i}) 
\end{equation}
This process ends until generating a special end-of-sequence token \texttt{<EOS>} or reaching a specified summary length $m$. 
We use the  Transformer-base\citep{vaswani2017transformer}  architecture as the backbone. 
For considering more contexts, we adopt the segment-level recurrence strategy, similar to \citet{dai-etal-2019-transformer}, where 
the hidden states computed
for far away snippets are fixed and cached to be reused for the next new snippet. 
Gradients are not propagated to these far away snippets for memory and computation efficiency.
This strategy allows the model to exploit information in history to the largest extent.  
\subsection{Expanding Summaries to Texts}
Next, we expand each summary $ \mathbf{s}^i$  to the full text for each segment by sequentially generating its constituent words
\begin{equation}
    p(\mathbf{y}^i|\mathbf{y}^{<i}, \mathbf{s}^i)  = \prod_{l \in [1,n_i]} p(y^i_l |\mathbf{y}^i_{<l}, \mathbf{s}^i, \mathbf{y}^{<i})
\end{equation}
which has the same termination conditions as in the summarization generation. 
\subsection{Training and Inference}
\paragraph{Training} For summary generation, the transformer model takes $[\mathbf{y}^{<i}; \mathbf{s}^{<i}]$ as the input and is optimized by minimizing the NLL loss $-\log p({\mathbf{s}}^i|\mathbf{y}^{<i}; \mathbf{s}^{<i})$. 
Due to the memory limitation, we limit $\mathbf{y}^{<i}$ to  preceding  384 tokens, and 
$\mathbf{s}^{<i}$ to 128 tokens at training. 
It is worth noting that the 384 tokens of $\mathbf{y}^{<i}$ mostly come from the segment right before, i.e., $\mathbf{y}^{i-1}$, while 
$\mathbf{s}^{<i}$  comes from multiple preceding segments since the summary is more concise.

For the summary expanding stage, the transformer model takes $[\mathbf{y}^{<i}; \mathbf{s}^{i} ]$ as input and 
is optimized by minimizing the NLL loss $-\log p(\hat{\mathbf{y}}^i| \mathbf{y}^{<i}; \mathbf{s}^{i})$.
The two models, i.e., the summary generation and the  summary expansion model share parameters, with a task-specific token appended to the start to notify the model on what to generate, summaries or segments.

%To compute the reconstruction score in \ref{extract}, we train a separate Transformer-base model. The model is trained on the same dataset as our main generation model does, and is optimized to minimize the reconstruction NLL loss in each segment: $-\sum_{\mathbf{y}^i\in\mathcal{D}}\sum_{\mathbf{y}^{i,j}}\log p(\mathbf{y}^i |\mathbf{y}^{i,j})$.

%%%%%%%%%%%%%%%%%%%%%%%%%%%%%%%%%%%%%%%%%%%%%%
%%%% xiaoya xiaoya xiaoya xiaoya 
%%%%%%%%%%%%%%%%%%%%%%%%%%%%%%%%%%%%%%%%%%%%%%
\paragraph{Inference} 
At test time, we first use beam search with beam size $5$ to generate summaries.
Given the generated summary, beam search is used again to generate the corresponding segment.
We consider more contexts at test time, where
$\mathbf{y}^{<i}$ is limited to  1,156 tokens and $\mathbf{s}^{<i}$ is limited to 512 tokens. 

Additionally, we augment the vanilla beam search with the strategy of mutual information reranking  \cite{li2015diversity,fang2015captions}.
The key point of mutual information is to,  
instead of merely handling the uni-directional dependency from the source to target based on the forward probability $\log p(\text{target}|\text{source})$, 
it
models the mutual dependency between the source and target in sequence-to-sequence generation, i.e., the combination of 
the forward probability $\log p(\text{target}|\text{source})$ and 
the backward probability $\log p(\text{source}|\text{target})$. 
Specifically in our case, during summary generation, $\mathbf{s}^i$ is generated as follows:
 \begin{equation}
 \begin{aligned}
 \mathbf{s}^i = \arg\max_{\mathbf{s}^i} & [ \log p(\mathbf{s}^i |\mathbf{y}^{<i}, \mathbf{s}^{<i})  + \\
 & \log p(\mathbf{s}^{i-1} |\mathbf{y}^{<i-1}, \mathbf{s}^{i}) ]
  \label{summary-mutual}
 \end{aligned}
 \end{equation}
 where $p(\mathbf{s}^{i-1} |\mathbf{y}^{<i}, \mathbf{s}^{i})$ is the backward probability of predicting the preceding summary $\mathbf{s}^{i-1}$ given $\mathbf{s}^{i}$. 
Since direct decoding from Eq.\ref{summary-mutual} is infeasible, we follow the practical solution in \citet{li2015diversity}, where we first generate an $N$-best list based on the forward probability 
$p(\mathbf{s}^i |\mathbf{y}^{<i}, \mathbf{s}^{<i})$,\footnote{We simplify $p(\mathbf{s}^{i-1} |\mathbf{y}^{<i-1}, \mathbf{s}^{i})$
   as $p(\mathbf{s}^{i-1} | \mathbf{s}^{i})$, where we train a seq2seq model to predict the preceding summary given the current summary.} and then rerank the $N$-best list by  combining the forward probability and the backward probability. 

Similar strategy can also be applied to the summary expanding stage, where  $\mathbf{y}^i$ is obtained as follows:
 \begin{equation}
 \begin{aligned}
\mathbf{y}^i= \arg\max_{\mathbf{y}^i} &[ \log p(\mathbf{y}^i|\mathbf{y}^{<i}, \mathbf{s}^{i})  \\
& + \log p(\mathbf{y}^{i-1}| \mathbf{y}^i) ]
  \label{mutual}
 \end{aligned}
 \end{equation}
The backward probability $ p(\mathbf{y}^{i-1}| \mathbf{y}^i)$ predicts the preceding  segment given the current segment. 
Again, beam search is combined with reranking to approximately find the optimal result. 

\subsection{Slicing Texts based on Coherence Scores}
One more thing we need to care about is how to slice the text into segments. 
The simplest way is to slice the full text equally. But this is sub-optimal since the break point could be in the middle of two closely related sentences and one segment might contain multiple aspects. 

We thus propose a slicing strategy  based on sentence-level  coherent scores. 
Using the Next Sentence Prediction (NSP) from BERT \citep{devlin2018bert}, we are able to measure the coherence score between two consecutive sentences with index $i$ and $i+1$, denoted by Score($i, i+1$).  Given a full text $\mathbf{y} = \{y(1), y(2), ..., y(T)\}$, let $T$ denote the number sentences in $\mathbf{y}$, and $y(i)$ denote the $i$th sentence. 
Given a fixed value $K$ for the number sliced segments, $\mathbf{y}$ will be sliced into $K$ segments, i.e.,  $\mathbf{y}^1, \mathbf{y}^2, ..., \mathbf{y}^K$, where each $\mathbf{y}^k$ consists of a group of consecutive sentences from $\mathbf{y}$. 
Let $G_k$ denote the list of  indexes of sentence in original $\mathbf{y}$,
where $G_k[1]$ denotes the index of the first sentence in $G_k$, $G_k[2]$ denotes the second sentence, etc. 
Let $R_k= |G_k|$ denote the number of sentences in $G_k$.

 We wish to maximize the coherence scores between two consecutive sentences within the same segment and minimize the score between two consecutive sentences belonging to different segments, giving the following objective to optimize:
\begin{equation}
\begin{aligned}
L = &\sum_{k=1}^{K}\sum_{i\in [1, R_k-1]}  \text{Score}(G(k)[i], G(k)[i+1]) \\
& - \sum_{k=1}^{K-1} \text{Score}(G[k][R_k], G[k][1])
\end{aligned}
\label{coherence}
\end{equation}
where $\text{Score}(G[k][R_k], G[k][1])$ the coherence score between the ending sentence of a segment and the starting sentence of the next segment. 
Given $\text{Score}(i,j)$, Eq.\ref{coherence} can be readily solved using linear programming. 

\section{Experiments}
\label{experiments}

\begin{table}[t]
  \centering
  \small
  \scalebox{0.9}{
  \begin{tabular}{lcccc}
    \toprule
     & \multicolumn{2}{c}{\textit{WikiText-103}} & \multicolumn{2}{c}{\textit{BookCorpus}}\\
    {\bf Model} & PPL$\downarrow$ & \# Para & PPL$\downarrow$ & \# Para \\\midrule 
    &\multicolumn{4}{c}{\textit{Base}} \\
    {\it Vanilla} & 25.0 & 130M & 29.0& 130M \\
    {\it WritingPrompts-Keyword} & 23.8 & 135M & 28.3 & 135M \\
    {\it WritingPrompts-Sentence} & 24.1 & 135M & 28.6& 135M \\ 
    {\it Progressive WritingPrompts} & 23.3 & 150M & 27.7&150M \\
    {\it SOE} & {\bf 22.2} & 132M & {\bf 25.7}& 132M\\
    \midrule
    &\multicolumn{4}{c}{\textit{Large}} \\
    {\it Vanilla} & 20.0 & 220M & 24.8 & 220M\\
    {\it SOE} & {\bf 17.4} & 224M & {\bf 22.5}& 224M \\\bottomrule
  \end{tabular}
  }
  \caption{Perplexity of different models on WikiText-103 and BookCorpus. {\it Vanilla} stands for our implementation of Transformer-XL \citep{dai-etal-2019-transformer}.}
  \label{results_ppl}
\end{table}

\begin{table*}[t]
  \centering
  \small
  \scalebox{0.95}{
  \begin{tabular}{lccccccc}
    \toprule
     & \multicolumn{3}{c}{\textit{MSJ}$\uparrow$} & \multicolumn{2}{c}{\textit{Diversity}$\uparrow$} & \multicolumn{1}{c}{\textit{Adversarial Success}$\uparrow$} & \multicolumn{1}{c}{\textit{S-Level Coherence}$\uparrow$}\\
    {\bf Model} & MSJ-2 & MSJ-3 & MSJ-4 & D-1 & D-2 & Adversarial Success & NSP \\\midrule
    & \multicolumn{7}{c}{\textit{Base}} \\
     {\it Vanilla}& 62.6& 41.5 & 16.9& 7.4& 19.8&0.037&0.812\\
     {\it WritingPrompts-Keyword} &63.0&42.2&17.5& 8.9&22.0&0.057&0.836\\
     {\it WritingPrompts-Sentence} &63.1&42.2&17.7&8.5&21.0&0.046&0.834\\
     {\it Progressive WritingPrompts}& 63.9&42.5&18.0&10.7&25.9&0.055&0.854\\
     {\it SOE} &64.8&43.9&19.4&16.4&34.3&0.072&0.870\\
     {\it SOE+MI} &{\bf 65.2}& {\bf 44.4}& {\bf 20.0}&{\bf 20.6}&{\bf 40.8}&{\bf 0.103}&{\bf 0.881}\\\bottomrule
  \end{tabular}
  }
  \caption{Results of different models in terms of diversity, adversarial success, MSJ and sentence-level coherence on the BookCorpus corpora. {\it Vanilla} stands for our implementation of Transformer-XL \citep{dai-etal-2019-transformer}. ``D-$n$'' stands for ``Distinct-$n (n=1,2)$'', and {\it MI} stands the results for mutual information 
  reranking. }
  \label{results_other}
\end{table*}

\begin{table}[t]
  \centering
  \small
    \scalebox{0.95}{
  \begin{tabular}{lcc}
    \toprule
    {\bf Model} & Distinct-1$\uparrow$ & Distinct-2$\uparrow$\\\midrule
    & \multicolumn{2}{c}{{\it Large}}\\
    {\it Vanilla} &  11.7 & 25.5 \\
    {\it SOE} &  24.1 & 45.0 \\
    {\it SOE+MI} & 29.3 & 48.8\\
    \bottomrule
  \end{tabular}
  }
  \caption{Results of different models with large volumes in terms of diversity on the BookCorpus dataset.}
  \label{results_large}
\end{table}

In this section, we present experiment results. 
For different methods to generate summaries, we find that the performance of 
{\it Reconstruction} consistently outperforms the rest in our preliminary results. 
We thus only report results from {\it Reconstruction} in the section. We will get back to analysis on different summary generation methods in the ablation study section. 

\subsection{Datasets and Evaluation Metrics}
We need a corpus of contiguous and long text to test SOE.
We use two word-level datasets, 
WikiText-103 \citep{merity2016pointer} and the BookCorpus dataset \citep{zhu2015aligning}. WikiText-103 contains 103M training words from 28K articles, with an average length of 3.6K words per article. 
WikiText-103 can be used to test the ability of modeling long-term dependencies.
The BookCorpus dataset is a more suitable dataset for our purpose, with much longer and more contiguous texts. 
It contains a total number of roughly 1 billion words and 74 million sentences from 11k books, with an average length of 89K words for each book. The average number of words per sentence is 13.
For both datasets, we predict the last 2,000 tokens at test time. 

We use Perplexity (PPL),  Diversity (Distinct-$n$) \citep{li-etal-2016-diversity}, Adversarial Success \citep{kannan2017adversarial,li2017adversarial}, MS-Jaccard (MSJ) \cite{montahaei2019jointly} and Sentence-Level Coherence \citep{tan2020progressive} as evaluation metrics. 
\begin{tightitemize}
    \item \textbf{Perplexity (PPL)} Perplexity measures how fluent a piece of generated text could be \citep{dai-etal-2019-transformer}. We use PPL as the basic evaluation metric in our experiments.
    \item \textbf{Diversity} Perplexity  cannot measure how diverse the generated text is. We thus use the scaled number of distinct unigrams (Distinct-1) and bigrams (Distinct-2) to demonstrate the degree of diversity \citep{li-etal-2016-diversity} for generated texts.
    \item \textbf{Adversarial Success}  Inspired by adversarial evaluations 
    \citep{bowman2016generating,kannan2017adversarial,li2017adversarial},
    we use the adversarial success metric, which is defined as the fraction of a
    model successfully fooling a trained evaluator to believe that machine-generated texts are from humans.
    The evaluator is a binary classification model. At the training time, it takes as inputs machine-generated texts and original texts, and are trained to discriminate them. 
    At test time, adversarial success is the value  $1- acc$, where $acc$ denotes the accuracy of the trained evaluator predicting machine-generated texts as machine-generated.
    Higher values of adversarial success denotes better text quality. 
    \item \textbf{MS-Jaccard (MSJ)}
    MSJ measures the similarity of the $n$-gram
    frequencies between the generated texts and the golden texts  \cite{montahaei2019jointly}. We report MSJ-2, -3 and -4.
    \item \textbf{Sentence-Level Coherence}
    PPL, MSJ and diversity scores do not reflect the sentence-level coherence of generated texts.
    We adopt the strategy in \citet{tan2020progressive} where 
    Next Sentence Prediction (NSP) from 
     pretrained BERT model \citep{devlin2018bert} is used as a metric to measure the coherence between each sentence and its next sentence. 
    We report average NSP scores for all consecutive sentence pairs within the generated text.
\end{tightitemize}

\subsection{Baselines}
In this paper, we use Transformer-XL~\citep{dai-etal-2019-transformer}, WritingPrompts~\citep{fan2018hierarchical}, and Progressive WritingPrompts~\citep{tan2020progressive} as baselines. 
More details of the baseline models can be found at Appendix \ref{sec:appendix}. 

\begin{comment}
\paragraph{Transformer-XL} 
Transformers with segment-level recurrence strategy \cite{dai-etal-2019-transformer} naturally constitutes a baseline. 
The model sequentially generates texts in a word-by-word fashion.  

\paragraph{WritingPrompts} 
 first predicts a list of keywords or a single prompt, and then generates the full text given the prompt \cite{fan2018hierarchical}.  Different from \citet{fan2018hierarchical}, where golden prompts for stories are available, we do not readily have the golden prompts. 
We thus use the extractive strategies described in Section \ref{extract}, i.e, the TF-IDF method to pick the keyword list as the prompt  (denoted by {\it WritingPrompts-keyword}) and the reconstruction method to select the highest ranking sentence as the prompt (denoted by {\it WritingPrompts-sentence}).

\paragraph{Progressive WritingPrompts} 
The progressive strategy proposed in \citet{tan2020progressive} which involves multiple stages of prompt generation. 
Each stage produces a more fine-grained sequence than the stage that comes before, and is used as the input to generate the prompt for the next stage.
We follow the protocols in \citet{tan2020progressive}
and use the 
TF-IDF score to obtain golden prompts for each stage. The number of stages is set to 4.
\end{comment}

For all models, we use Adam \citep{kingma2014adam} with learning rate of 1e-4, $\beta_1$ = 0.9,
$\beta_2$ = 0.999, 
rate warmup over the first 10,000 steps, and linear
decay of the learning rate. We use a dropout rate of 0.1 on all layers.
 
\subsection{Results}
Table \ref{results_ppl} shows the results of perplexity for different models on the WikiText-103 and BookCorpus datasets. On both datasets, SOE achieves the lowest PPL compared to  baselines Transformer-XL \citep{dai-etal-2019-transformer}, WritingPrompts \citep{fan2018hierarchical} and Progressive \citep{tan2020progressive}. In particular, for WikiText-103, we gain a PPL decrease -2.8, -1.6 and -1.1 against our implemented Transformer-XL, WritingPrompts-Sentence and Progressive, while having the same or even fewer parameters. 
Similar trend can be observed on BookCorpus.

Table \ref{results_other} and Table \ref{results_large} show the results for MSJ, diversity, adversarial success and sentence-level coherence scores. 
As can be seen, WritingPrompt based models generally outperform the Transformer-XL model, which adopts the word-by-word generation strategy.  This  validates the superiority of two-step
 generation strategy over the naive word-by-word generation strategy for long-text generation. 
The progressive WritingPrompt model, which involves multi-step of generation and expanding, outperforms the 
one-step  the WritingPrompt-keyword and WritingPrompt-sentence model, which is in accord with our expectation. 
SOE  achieves significantly better results compared to Vanilla, WritingPrompts and Progressive models in terms of all evaluation metrics, showing that
the proposed method can produce more fluent, coherent and diverse texts.
The consistent performance boosts on all metrics 
demonstrate the importance of modeling discourse-level dependencies and 
necessity of summary expanding strategy for long-text generation. 

Additionally, we observe additional performance boosts by mutual information (MI), especially for diversity and adversarial success. This is in accord with our expectation: since mutual information is able to build bidirectional dependencies between the source and the target,  models enhanced with mutual information can generate better summaries, and the phenomenon of generic and repetitive generation can be alleviated \citep{li-etal-2016-diversity}, leading to more diverse results.

%For large models in Table \ref{results_large}, we only report the diversity results. Combining the results in Table \ref{results_other}, it shows that the two-stage summary-expanding strategy is beneficial for generating more diverse texts, and avoids being trapped in excessive local word predictions.

\section{Ablation Studies}
\label{ablations}
\subsection{The Effect of Segment Length}
\begin{figure}[t]
  \centering
  \includegraphics[scale=0.45]{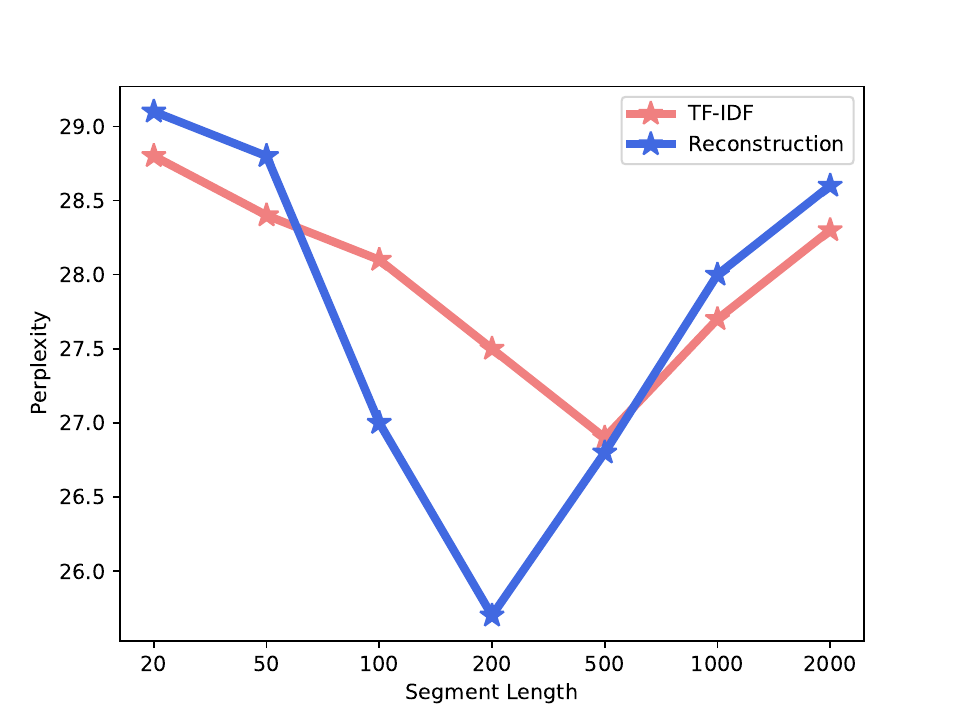}
  \caption{PPL on the BookCorpus dataset w.r.t. different segment lengths.}
  \label{segment_length}
\end{figure}

The size of the segment can be neither too big nor too small: extremely long segments, 
might contain too many aspects or topics for the summary  to summarize, in which case 
 the model will degenerate into the WritingPrompts model \cite{fan2018hierarchical}. 
For too short segments, the summary cannot provide high-level guidance. 
We thus need to find the sweet spot for the segment length. 
Figure \ref{segment_length} shows results on the BookCorpus dataset. 
It is clear from the figure that  too short segments and too long segments both  lead to inferior performances. 
%For both TF-IDF and Reconstruction methods, the trend is the same: the PPL first drops as we increase the segment length from 20 to 200 for Reconstruction and 500 for TF-IDF, and then the PPL keeps rising as we continue increasing the length. 

\subsection{The Effect of Summary Generation Strategies}
It is worthwhile to  explore how different summary extraction methods affect the final performances. To this end, we conduct experiments on the BookCorpus dataset, using different summary extraction methods, i.e., Random, TextRank, TF-IDF and Reconstruction. 
Table \ref{summary_generation} shows the results. 
We first compare the ppl for summary generation, where the reconstruction model achieves the lowest ppl and thus produces summaries that are the easiest to predict given preceding contexts. 
It is also interesting to see that across all summary generation strategies, ppl for summarization generation is significantly larger than text prediction, which is reasonable since (1) generating summaries for the upcoming segment requires more generalization abilities; and (2) there are more diverse options for what the next segment should talk about than the local choices for what the next sentence should talk about. 
For the final text-generation ppl,
reconstruction achieves the best results, in terms of PPL, MJ-3 and MJ-4. TextRank and TF-IDF are better than Vanilla.
Interestingly, the strategy of using random sentences as summaries performs worse than without  summaries, which can be explained by providing no  guidance is better than incorrect guidance.

\begin{table}[t]
  \centering
  \small
    \scalebox{0.95}{
  \begin{tabular}{lccc}
    \toprule 
    {\bf Method}&Summary PPL$\downarrow$ &Text PPL$\downarrow$& MJ-4 $\uparrow$\\\midrule 
    {\it Vanilla}&- & 29.0  & 16.9\\
    {\it Random}& 40.1& 30.2  & 15.5 \\
    {\it TextRank}&30.7 & 26.2  & 17.8\\
    {\it TF-IDF}&33.0& 26.9   &17.3 \\
    {\it Reconstruction}&{\bf 30.4} & {\bf 25.7}  & {\bf 19.4}\\\bottomrule
  \end{tabular}
  }
  \caption{Performances of different summary extraction methods described in Section \ref{extract}. {\it Vanilla} is the plain model that generates tokens one by one without summaries.}
  \label{summary_generation}
\end{table}

\subsection{The Effect of Coherence-based Text Slicing}
We replace the coherence-based text slicing strategy with the naive equal slicing strategy, and see how this 
will negatively affect the performance. On the BookCorpus dataset, we observe an increase of summary generation ppl from 30.4 to 30.9, and 
an +0.7 increase of PPL from  25.7 to 26.4 in token generation, which demonstrates the importance of slicing text into coherent segments for generation.
But it is also worth-noting that, even with the native equal slicing strategy, SOE still performs significantly better than other baseline models.
\subsection{Decoupling The  Effects of Summaries}
The positive effects from summaries are two-fold:
(1) it provides  high-level guidance for segment generation; and (2)
with far-away segments being concisely represented by summaries, it gives the model the ability to consider longer contexts. 
To quantitatively  measure the influences from both aspects, we conduct the following experiments: 
at test time, 
for the computation $p({\mathbf{s}}^i|\mathbf{y}^{<i}; \mathbf{s}^{<i})$ and $ p({\mathbf{y}}^i| \mathbf{y}^{<i}; \mathbf{s}^{i} )$, 
the model can only access summaries  for segments that are used as contexts.
In other words, only summaries within the 1,156 tokens of preceding contexts can be fed as inputs. 
This is different from the original version of SOE, in which  $\mathbf{s}$ can extend to preceding contexts until the limit of 512 tokens is reached. 
We did not retrain the model, but add this limitation at test time.
On the BookCorpus dataset, this leads to an increase of 0.8 in PPL (25.7 vs 26.5), and a decrease of 0.5 and 0.8 in MJ-3 (43.5 vs 43.9) and MJ-4 (18.6 vs 19.4).  

\subsection{Simplifying  $p({\mathbf{s}}^i|\mathbf{y}^{<i}; \mathbf{s}^{<i})$} 
Here we explore different simplifications for  $p({\mathbf{s}}^i|\mathbf{y}^{<i}; \mathbf{s}^{<i})$. 
For $p({\mathbf{s}}^i|\mathbf{y}^{<i}; \mathbf{s}^{<i})$, the current summary is generated based on both previous summaries and segment tokens. 
We can simplify it as $p({\mathbf{s}}^i |\mathbf{s}^{<i})$, where previous segment tokens are not fed as inputs to predict the summary, which will significantly decreases computing complexity. 
On the BookCorpus dataset, we observe an increase of PPL in summary generation from 30.4 to 31.2, 
which subsequently leads to an +0.9 increase of PPL from  25.7 to 26.6 in token generation.

\subsection{Convergence Speed}
\begin{figure}[t]
  \centering
  \includegraphics[scale=0.45]{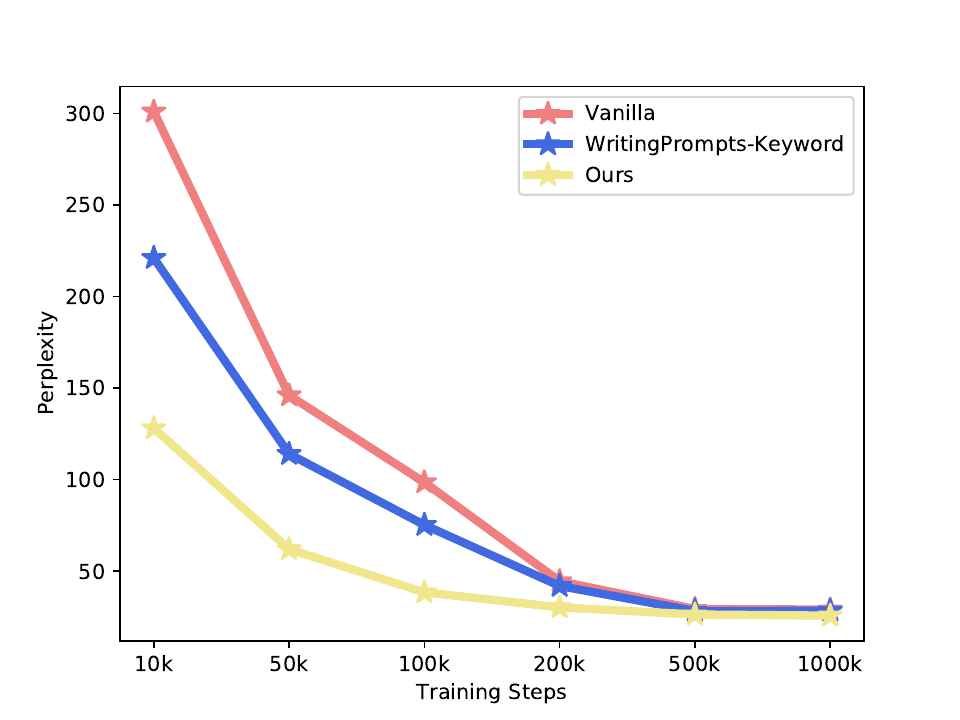}
  \caption{Convergence speed for different models.}
  \label{convergence}
\end{figure}

At last, we investigate how quickly different models converge. Results are shown in Figure \ref{convergence}. With the guidance of extracted summaries, SOE has a conspicuously faster convergence speed, where at about 200K training steps it has approximately reached the best result while the other two models --- Vanilla and WritingPrompts --- do not converge until 1000K training steps. The WritingPrompts model converges faster than then Vanilla because of the high-level guidance from prompts. 

\section{Conclusion}
\label{conclusion}
In this paper, 
we propose a 
two-step 
hierarchical generation strategy for long-text generation: 
the model first generates the summary for each segment conditioning on previous summaries, and  
next, 
 each summary is expanded to form the full text segment. 
The proposed strategy provides  high-level guidance for local text generation, and enables high-level discourse dependencies to be captured. 
  Extensive experiments demonstrate that SOE produces long texts with significantly better quality,

% Use \bibliography{yourbibfile} instead or the References section will not appear in your paper

\section*{Acknowledgement}
We would like to thank anonymous reviewers for their comments and suggestions.
This work is supported by the Key R \& D Projects of the Ministry of Science and Technology (2020YFC0832500), Program of Zhejiang Province Science and Technology (2022C01044).

\bibliography{custom}
\bibliographystyle{acl_natbib}

\appendix
\section{Methods for Selecting Summary Sentences}
\label{sec:appendix0}
We investigate Random, TF-IDF~\citep{Ramos2003UsingTT}, TextRank~\citep{mihalcea2004textrank} methods to access the importance of selecting summary sentences. 

 \paragraph{Random} For comparing purposes, we use a random sentence as the summary.
 \paragraph{TF-IDF} We take the sentence with the highest average TF-IDF score \citep{Ramos2003UsingTT} as the golden summary ${\mathbf{s}}^i$. A word is assigned a score by TF-IDF that scales proportionally to the number of times the word appears in the document and is offset by the number of documents in the corpus that contain the word, which can be expressed as $N_w\cdot\log(\frac{N_d}{N_{dw}})$, where $N_w$ is the word count, $N_d$ is the total number of documents and $N_{dw}$ is the total number of documents containing the word.
 \paragraph{TextRank} TextRank \citep{mihalcea2004textrank} is a weighted graph with tokens as nodes and the similarity between nodes as edges. We use BERT \citep{devlin2018bert} to compute the similarities between sentences and then rank them based on the TextRank algorithm.

\section{Model Baselines}
\label{sec:appendix}
In this paper, we use Transformer-XL, WritingPrompts, and Progressive WritingPrompts as baselines. 

\paragraph{Transformer-XL} 
Transformers with segment-level recurrence strategy \cite{dai-etal-2019-transformer} naturally constitutes a baseline. 
The model sequentially generates texts in a word-by-word fashion.  

\paragraph{WritingPrompts} 
 first predicts a list of keywords or a single prompt, and then generates the full text given the prompt \cite{fan2018hierarchical}.  Different from \citet{fan2018hierarchical}, where golden prompts for stories are available, we do not readily have the golden prompts. 
We thus use the extractive strategies described in Section \ref{extract}, i.e, the TF-IDF method to pick the keyword list as the prompt  (denoted by {\it WritingPrompts-keyword}) and the reconstruction method to select the highest ranking sentence as the prompt (denoted by {\it WritingPrompts-sentence}).

\paragraph{Progressive WritingPrompts} 
The progressive strategy proposed in \citet{tan2020progressive} which involves multiple stages of prompt generation. 
Each stage produces a more fine-grained sequence than the stage that comes before, and is used as the input to generate the prompt for the next stage.
We follow the protocols in \citet{tan2020progressive}
and use the 
TF-IDF score to obtain golden prompts for each stage. The number of stages is set to 4.

%This is an appendix.

\end{document}